\UseRawInputEncoding
\documentclass[journal]{IEEEtran}

\usepackage{cite}
\usepackage{amsmath,amssymb,amsfonts}
\usepackage{graphicx}
\usepackage{textcomp}
\usepackage{algorithmic}
\usepackage{algorithm}
\usepackage{color,soul}

\usepackage{fancyhdr}
\usepackage{tikz}

\def\BibTeX{{\rm B\kern-.05em{\sc i\kern-.025em b}\kern-.08em
    T\kern-.1667em\lower.7ex\hbox{E}\kern-.125emX}}
\begin{document}

\title{\textit{FourierNet}: Shape-Preserving Network for Henle's Fiber Layer Segmentation in Optical Coherence Tomography Images}
\author{Selahattin Cansiz, Cem Kesim, Sevval Nur Bektas, Zeynep Kulali, Murat Hasanreisoglu, and \\Cigdem Gunduz-Demir, \IEEEmembership{Member, IEEE}
\thanks{This work was partly supported by the Scientific and Technological Research Council of Turkey, project no: T{\"U}B\.{I}TAK 120E497.}
\thanks{S. Cansiz is with the Department of Computer Engineering, Bilkent University, 06800 Ankara, Turkey (e-mail: selahattin@bilkent.edu.tr).}
\thanks{C. Kesim is with the Department of Ophthalmology, Koc University School of Medicine, 34010 Istanbul, Turkey (e-mail: ckesim@kuh.ku.edu.tr).}
\thanks{S. N. Bektas and Z. Kulali are with the Koc University School of Medicine, 34010 Istanbul, Turkey (e-mail: sbektas19@ku.edu.tr; zkulali15@ku.edu.tr).}
\thanks{M. Hasanreisoglu is with the Department of Ophthalmology, Koc University School of Medicine, and Koc University Research Center for Translational Medicine, 34010 Istanbul, Turkey (e-mail: mhasanreisoglu@ku.edu.tr).}
\thanks{C. Gunduz-Demir is with the Department of Computer Engineering and KUIS AI Center, Koc University, 34450 Istanbul, Turkey and  with the Department of Computer Engineering, Bilkent University, 06800 Ankara, Turkey(e-mail: cgunduz@ku.edu.tr; gunduz@cs.bilkent.edu.tr).}

}

\maketitle

\begin{abstract}
The Henle's fiber layer (HFL) in the retina carries valuable information on the macular condition of an eye. However, in the common practice, this layer is not separately segmented but rather included in the outer nuclear layer since it is difficult to perceive HFL contours on standard optical coherence tomography (OCT) imaging. Due to its variable reflectivity under an imaging beam, delineating the HFL contours necessitates directional OCT, which requires additional imaging. This paper addresses this issue by introducing a shape-preserving network, \emph{FourierNet}, that achieves HFL segmentation in standard OCT scans with the target performance obtained when directional OCT scans are used. \emph{FourierNet} is a new cascaded network design that puts forward the idea of benefiting the shape prior of HFL in the network training. This design proposes to represent the shape prior by extracting Fourier descriptors on the HFL contours and defining an additional regression task of learning these descriptors. It then formulates HFL segmentation as concurrent learning of regression and classification tasks, in which Fourier descriptors are estimated from an input image to encode the shape prior and used together with the input image to construct the HFL segmentation map. Our experiments on 1470 images of 30 OCT scans reveal that quantifying the HFL shape with Fourier descriptors and concurrently learning them with the main task of HFL segmentation lead to better results. This indicates the effectiveness of designing a shape-preserving network to improve HFL segmentation by reducing the need to perform directional OCT imaging.
\end{abstract}


\thispagestyle{fancy}

\chead{This work has been submitted to the IEEE for possible publication. Copyright may be \\transferred without notice, after which this version may no longer be accessible.}
\renewcommand{\headrulewidth}{0.0pt}

\section{Introduction}
\label{sec:introduction}
\IEEEPARstart{O}{ptical} coherence tomography (OCT) is an essential retinal imaging equipment that allows the visualization of individual layers of the retina. Ophthalmologists employ OCT scans to diagnose eye-related diseases and understand their severity. Within the retina layers, the layer composed of photoreceptor axons, known as the Henle's fiber layer (HFL), provides significant information and the changes in its thickness are commonly associated with the macular condition in diseased retinas~\cite{vannasdale2019henle}. However, due to its variable reflectivity under an imaging beam, it is challenging to separately segment HFL on standard OCT scans. Directional OCT, which obtains images by altering the entry position of the imaging beam at the pupil, emerges as an important technique for HFL segmentation~\cite{lujan2011revealing}. However, this technique is not a routine clinical procedure mostly because it necessitates significant amount of additional examination time, and thus, in the common practice, HFL is considered as a part of the outer nuclear layer. Therefore, there exists automatic HFL segmentation neither in commercially available OCT softwares~\cite{tian2016performance} nor in scholarly studies~\cite{fang2017automatic, pekala2019deep}. In response to this issue, this paper introduces a new cascaded neural network, which we call \emph{FourierNet}, for HFL segmentation in standard OCT scans. The proposed \emph{FourierNet} model achieves the performance of its counterpart, which uses scans obtained from the directional OCT technique as its inputs, but using only the standard OCT scans without requiring any non-routinely used imaging modalities. 

In order to facilitate HFL segmentation, the \emph{FourierNet} model proposes to employ prior knowledge on the shape of HFL in the network training. It proposes to quantify this prior shape knowledge with a function defined on the HFL contours. To this end, it expands the function in a Fourier series, uses the harmonic amplitudes of its Fourier coefficients as the Fourier descriptors of HFL, and represents this prior shape knowledge in the network design by defining a regression task of learning these Fourier descriptors. It then formulates HFL segmentation as concurrent learning of regression and classification tasks, in which Fourier descriptor maps are estimated from an input image to represent the prior shape knowledge on HFL and used along with the input image to estimate the segmentation label for each pixel in the image. \textit{FourierNet} achieves this learning by designing a cascaded fully convolutional network (FCN) that consists of an intermediate regression and a final classification task.

The contributions of this paper are three-fold:

\begin{itemize}
\item To the best of our knowledge, this is the first paper that has automatically segmented the Henle's fiber layer (HFL) in standard OCT scans. Although layer-wise segmentation of retinal OCT scans has been studied widely, none of the studies segment HFL separately but include it in the outer nuclear layer (ONL).
\item \emph{FourierNet} presents a cascaded FCN design where the intermediate task of Fourier descriptor estimation contributes to the final segmentation task by providing it with the shape-related information about what it needs to estimate. Since the network weights for these two tasks are updated at the same time, by minimizing a joint loss function, this concurrent learning more likely imposes the shape on the network. This, in turn, leads to better preserving the HFL shape, and thus, enhances HFL segmentation.
\item \emph{FourierNet} quantifies the HFL shape with a set of Fourier descriptors and devises a cascaded FCN design for their estimation. These Fourier descriptors, which were first suggested by Cosgriff in 1960~\cite{cosgriff60}, were also used in previous studies~\cite{zhang03,larsson11,zahn72}. However, the previous studies used these Fourier descriptors to characterize objects in different applications (e.g., object retrieval and recognition), but did not employ them in designing a shape-preserving FCN. On the other hand, \emph{FourierNet} uses the Fourier descriptors to devise a shape-preserving FCN for the first time, and demonstrates that this use is effective for more accurate HFL segmentation.
\end{itemize}

\section{Related Work}

In the literature, there have been many studies proposed for layer-wise segmentation of the retina. Earlier studies typically segment the retina layers by first identifying their boundary pixels and then refine them with optimization methods. These initial pixels are identified by either edge detection~\cite{mayer2010retinal} or training a classifier~\cite{mcdonough2015neural,fuller2007segmentation,chen2012three,lang2013retinal}. It is common to use graph-based optimization algorithms for pixel refinement. In~\cite{chen2012three},  it is proposed to apply a graph cut algorithm with probability constraints on the initial pixels segmented by a k-nearest neighbor classifier. Likewise, a final segmentation map is obtained by first applying soft constraints to utilize prior information from a learned model and then regularizing the distances between two segmented layers by a graph based algorithm~\cite{dufour2012graph}. In~\cite{chiu2015kernel}, a graph-cut algorithm and dynamic programming are used together to refine initial layers classified by kernel regression. Other refinement methods have also been used. In~\cite{mayer2010retinal}, initial pixels identified by edge detection are improved by minimizing an energy term that considers vertical gradients and regional smoothness. In~\cite{novosel2015loosely}, the predefined order of layers and thickness priors are employed to refine the segmented retinal layers.

More recent studies have widely used deep learning models, which remarkably improve layer-wise retina segmentation. Many studies use a U-Net~\cite{ronneberger2015u} based network, which contains symmetric encoder and decoder blocks~\cite{roy2017relaynet,he2017towards}. The literature also contains modified U-Net architectures. In~\cite{gu2019net}, a modified U-Net with a context extractor module, which generates high-level semantic feature maps, is used. In~\cite{apostolopoulos2017pathological}, retina layers are segmented designing an asymmetric U-shape network that combines residual building blocks with dilated convolutions. There also exist studies that combine deep learning models with post-processing techniques to correct pixel-wise classifications, and hence, obtain more accurate layers. For instance, a graph search algorithm is applied on the posterior probability maps outputted by a convolution neural network to find the final boundaries~\cite{fang2017automatic}. In another study~\cite{pekala2019deep}, the DenseNet architecture is combined with a Gaussian process regression to smooth the segmentation results. In~\cite{liu2018automated}, a random forest classifier is trained on handcrafted features along with deep features learned by a deep residual network and this trained classifier is used to obtain the contour probabilities of each retinal layer.  
 
Although all these studies yield promising results on the segmentation of retinal layers, none of them segment HFL separately. Additionally, they neither define Fourier descriptors to represent the shape prior of a retina layer nor integrate this information with the design of a neural network. On the other hand, the proposed \emph{FourierNet} model introduces a shape-preserving network for HFL segmentation for the first time and presents an effective way of representing the shape priors of HFL defining the Fourier descriptors on the contours of HFL.

\section{Methodology}
The proposed \emph{FourierNet} model relies on characterizing the HFL shape with Fourier descriptors (Sec.~\ref{sec:fd}), defining a regression task of estimating the maps of these Fourier descriptors (Sec.~\ref{sec:map}), and learning this regression task together with the main task of HFL segmentation by designing a cascaded FCN (Sec.~\ref{sec:fcn}). The following subsections give the details. The implementation is available at mysite.ku.edu.tr/cgunduz/downloads/FourierNet.

\subsection{Fourier Descriptors}
\label{sec:fd}
This work quantifies the shape of HFL with a function defined along its contour points. This is the distance-to-center function that outputs the distance from an object centroid to the boundary point at a given arc length. This function characterizes how subsequent boundary points distribute over the space to form up the object's contour, and thus, the shape of the object. For instance, it is a constant function for a circular object since the distance from any boundary point to the centroid (radius) is the same.

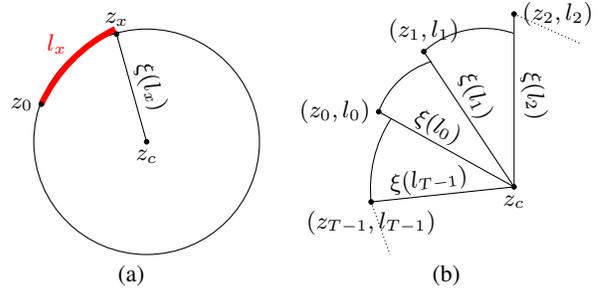
\begin{figure}
\centering
\small{
\begin{tabular}{cc}

\begin{tikzpicture}
\coordinate (c) at (-3,0);
\coordinate (v0) at (-4.4,0.50);
\coordinate (v1) at (-3.4,1.4330);
\coordinate (e1) at (-3.10,1.15);
\coordinate (e2) at (-3.95,1.3);
\draw[fill] (c) circle [radius=0.03];
\draw (c) circle [radius=1.5];
\draw[line width=0.8mm, red] (v0) arc (155:116:2.1cm);
\draw[fill] (v0) circle [radius=0.03];
\draw[fill] (v1) circle [radius=0.03];
\draw (c) -- (v1);
\node [below] at (c) {$z_c$};
\node [left] at (v0) {$z_0$};	
\node [above] at (v1) {$z_x$};	
\node [right, rotate=-70] at (e1) {$\xi(l_x)$};
\node [red, left] at (e2) {$l_x$};
\end{tikzpicture}
&
\begin{tikzpicture}
\coordinate (c) at (0.2,0.0);		\coordinate (v0) at (-1.6,1.0);		\coordinate (v1) at (-1.0,1.8);		\coordinate (v2) at (0.2,2.3);		\coordinate (v3) at (-1.7,-0.2);	
\coordinate (v4) at (1.1,1.9);		\coordinate (v5) at (-1.45,-0.9);		\coordinate (e0) at (-1.2,1.0);		\coordinate (e1) at (-0.5,1.6);		\coordinate (e2) at (0.4,1.65);		
\coordinate (e3) at (-0.9,-0.18);
\draw[fill] (c) circle [radius=0.03];	\draw[fill] (v0) circle [radius=0.03];	\draw[fill] (v1) circle [radius=0.03];	\draw[fill] (v2) circle [radius=0.03];	\draw[fill] (v3) circle [radius=0.03];
\draw[densely dotted] (v2) -- (v4);	\draw[densely dotted] (v3) -- (v5);	\draw (v0) arc (160:107:1.09cm);	\draw (v1) arc (132:71:1.20cm);	\draw (v3) arc (186:147:1.71cm);
\draw (c) -- (v0);				\draw (c) -- (v1);				\draw (c) -- (v2);				\draw (c) -- (v3);
\node [below] at (c) {$z_c$};				\node [left] at (v0) {$(z_0, l_0)$};					\node [below] at (v3) {$(z_{T-1}, l_{T-1})$};			
\node [above] at (v1) {$(z_1, l_1)$};			\node [right] at (v2) {$(z_2, l_2)$};		
\node [right, rotate=-28] at (e0) {$\xi(l_0)$};	\node [right, rotate=-65] at (e1) {$\xi(l_1)$};	\node [right, rotate=-82] at (e2) {$\xi(l_2)$};	\node [above, rotate=7] at (e3) {$\xi(l_{T-1})$};
\end{tikzpicture}
\\	(a) & (b)
\end{tabular}
}
\caption{(a) The distance-to-center function $\xi(.)$ defined on the domain of length $l_x \in [0,L]$ where $l_x$ denotes the arc length of a section of the continuous curve $\gamma$ from its starting point $z_0$ to the point $z_x$ of the same curve. (b) Illustration of a circular arc interpolation between the discrete points (pixels) of HFL contour $\gamma_h$. This interpolation will be used to define a continuous curve for Fourier descriptor calculations. In this illustration, consecutive pixels are drawn too separated from each other for demonstration purposes and each pixel is denoted with its coordinate $z_x$ and its corresponding arc length $l_x$.}
\label{fig:interpolation}
\end{figure}

Let $\gamma$ be a closed continuous curve with a length of $L$. Fourier descriptors are calculated for the distance-to-center function $\xi(.)$ on the domain of length $l_x \in [0,L]$ where $l_x$ denotes the arc length of  a section of the curve $\gamma$ from its starting point $z_0$ to the point $z_x$ of the same curve (Fig.~\ref{fig:interpolation}a). Since the contour of an object in a digital image does not form a continuous curve but contains finitely many discrete points (pixels), we assume an arc interpolation between these discrete points to define a continuous curve. Thus, the curve $\gamma_h$ that corresponds to the contour of HFL $h$ becomes the interpolation of $T$ boundary pixels, $\{z_0, z_1, ..., z_{T-1}\}$, each of which lies on the curve $\gamma_h$ at arc length $l_t$ (Fig.~\ref{fig:interpolation}b). 

The distance-to-center function $\xi(l_x)$ outputs the distance from the centroid $z_c$ to the point $z_x$ for which the arc length is $l_x$. This function is expanded in a Fourier series as
\begin{eqnarray}
\xi(l_x) = a_0 + \sum_{n=1}^{\infty} \left[a_n \cos\left(\frac{2\pi n l_x}{L} \right) + b_n \sin\left(\frac{2\pi n l_x}{L}\right)\right]
\end{eqnarray}
where
\begin{eqnarray}
\label{ak}
a_n &=& \frac{2}{L} \int_{0}^{L} \xi(l_x) \cos\left(\frac{2\pi n l_x}{L}\right) dl_x \\ 
\label{bk}
b_n &=& \frac{2}{L} \int_{0}^{L} \xi(l_x) \sin\left(\frac{2\pi n l_x}{L}\right) dl_x
\end{eqnarray}
For the curve $\gamma_h$, which is an interpolation of $T$ discrete pixels, Eqn.~\ref{ak} can be divided into $T$ intervals of $[l_{t-1},l_{t})$. Thus, the coefficient $a_n$ can be written as
\begin{eqnarray}
a_n = \frac{2}{L} \sum_{t=1}^{T} \int_{l_{t-1}}^{l_t} \xi(l_x) \cos\left(\frac{2\pi n l_x}{L}\right) dl_x
\end{eqnarray}
Here we use a circular arc interpolation to estimate the value of $\xi(l_x)$ for all lengths $l_x$ that do not correspond to any boundary pixels. Since the distance from the circle centroid to any point on a circle is the same, this interpolation gives $\xi(l_x)=\xi(l_{t-1}), \forall l_x \in [l_{t-1},l_t)$, and as a result, allows taking $\xi(l_x)$ outside the integral.
\begin{align*}
a_n =& \frac{2}{L} \sum_{t=1}^{T} \xi(l_{t-1}) \int_{l_{t-1}}^{l_t} \cos\left(\frac{2\pi n l_x}{L}\right) dl_x	\\
a_n =& \frac{1}{\pi n}  \sum_{t=1}^{T} \xi(l_{t-1}) \left[ \sin\left(\frac{2\pi n l_t}{L}\right) - \sin\left(\frac{2\pi n l_{t-1}}{L}\right) \right]	\\
a_n =& \frac{1}{\pi n}  \left[ \xi(l_0) \sin\left(\frac{2\pi n l_1}{L}\right) - \xi(l_0) \sin\left(\frac{2\pi n l_0}{L}\right) \right.	\\
&~~+ \xi(l_1) \sin\left(\frac{2\pi n l_2}{L}\right) - \xi(l_1) \sin\left(\frac{2\pi n l_1}{L}\right)	\\
&~~\vdots	\\
&\left.~~+ \xi(l_{T-1}) \sin\left(\frac{2\pi n l_T}{L}\right) - \xi(l_T) \sin\left(\frac{2\pi n l_{T-1}}{L}\right) \right]	
\label{eqn:ak-derivation}
\end{align*}
Since $\gamma_h$ is a closed curve, the last point $z_T$ is indeed the starting point $z_0$, and thus, $\xi(l_0)=\xi(l_T)$ and $\sin(\frac{2\pi n l_0}{L}) = \sin(\frac{2\pi n l_T}{L})$. By defining $\Delta \xi_t=\xi(l_{t-1})-\xi(l_t)$
\begin{eqnarray}
a_n = \frac{1}{\pi n} \sum_{t=1}^{T}   \Delta \xi_t~\sin\left(\frac{2\pi n l_{t}}{L}\right)
\end{eqnarray}
Following similar steps, the coefficient $b_n$ is expressed as:
\begin{eqnarray}
b_n = -\frac{1}{\pi n} \sum_{t=1}^{T}   \Delta \xi_t~\cos\left(\frac{2\pi n l_{t}}{L}\right)
\end{eqnarray}
For the $n$-th Fourier coefficients $(a_n, b_n)$, the polar coordinates are $(A_n, \alpha_n)$ where $A_n = \sqrt{{a_n}^2+{b_n}^2}$ is the harmonic amplitude and $\alpha_n = \arctan{ (b_n / a_n)}$ is the harmonic phase. 

This work uses the first $N$ harmonic amplitudes of a truncated expansion of $\xi(l_x)$ as a set of Fourier descriptors $\texttt{FD}(\gamma_h) = [A_1, A_2, ..., A_N]$ to characterize the contour $\gamma_h$ of a given HFL $h$, and hence, its shape. Note that when $N\to\infty $, the curve can be reconstructed using these harmonic amplitudes together with their corresponding harmonic phases. However, we do not use the harmonic phases to define a descriptor set as they provide less shape related information~\cite{zahn72}.

%

\begin{figure}
\centering
\small{
\begin{tabular}{cc}
\includegraphics[width=0.43\columnwidth]{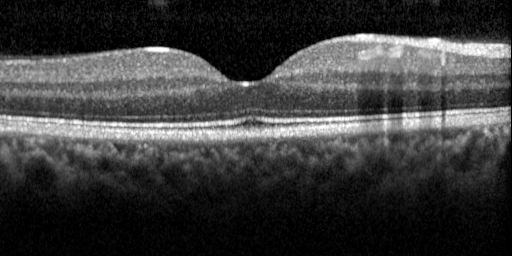} &
\includegraphics[width=0.43\columnwidth]{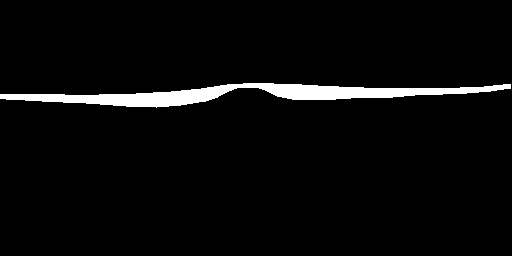} \\ 
(a) & (b) \\ 
\end{tabular}
\begin{tabular}{c}
\includegraphics[width=0.91\columnwidth]{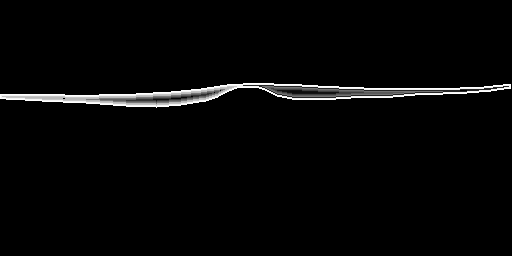} \\
(c) \vspace{-0.2cm} 
\end{tabular}
}
\caption{(a) An image taken from the standard OCT, (b) its manual HFL annotations, (c) the map of the first Fourier descriptors calculated with respect to the given HFL annotations.}
\label{fig:maps}
\end{figure}

\begin{figure*}
\begin{center}
\centerline{\includegraphics[width=17.5cm]{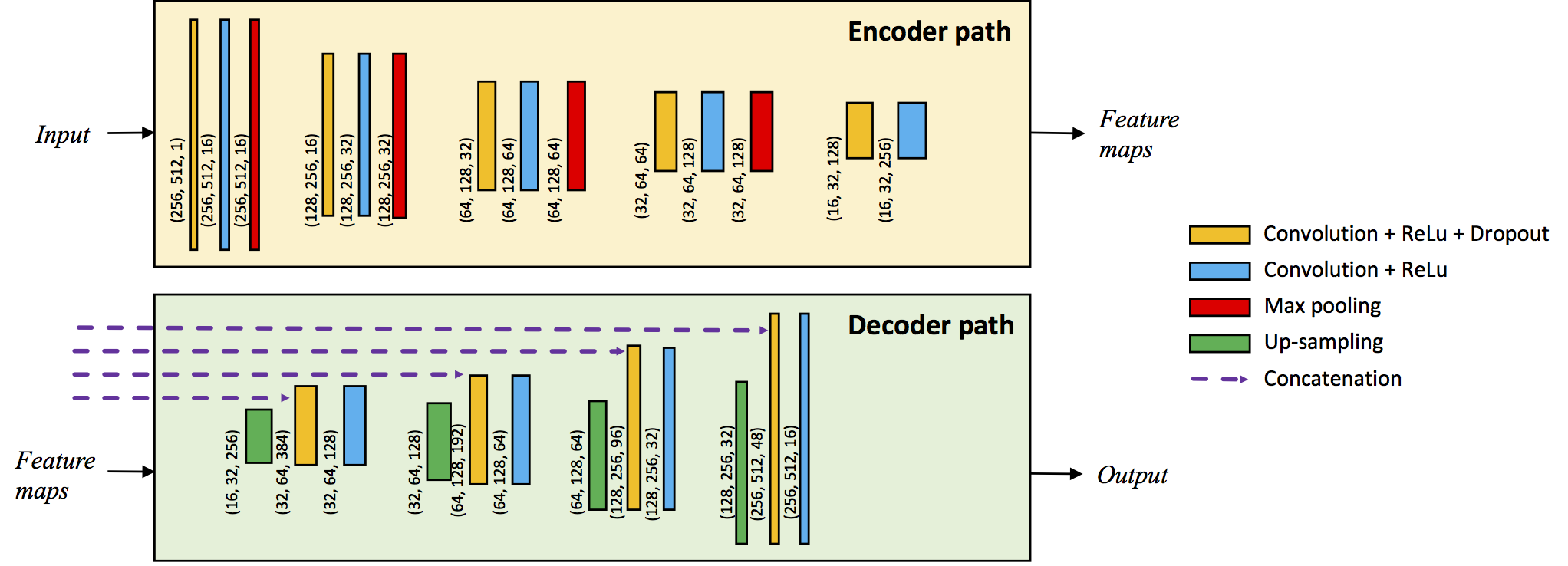}}
\vspace{-0.8cm}
\end{center}
\caption{Encoder and decoder paths used in the U-Net architectures. Each box corresponds to an operation, which is distinguishable by its color. The input to each operation is a multi-channel map with its dimensions and number of channels being indicated in order on the left side of the box.}
\label{fig:base-model}
\end{figure*}

\begin{figure*}
\begin{center}
\vspace{0.6cm}
\centerline{\includegraphics[width=17.5cm]{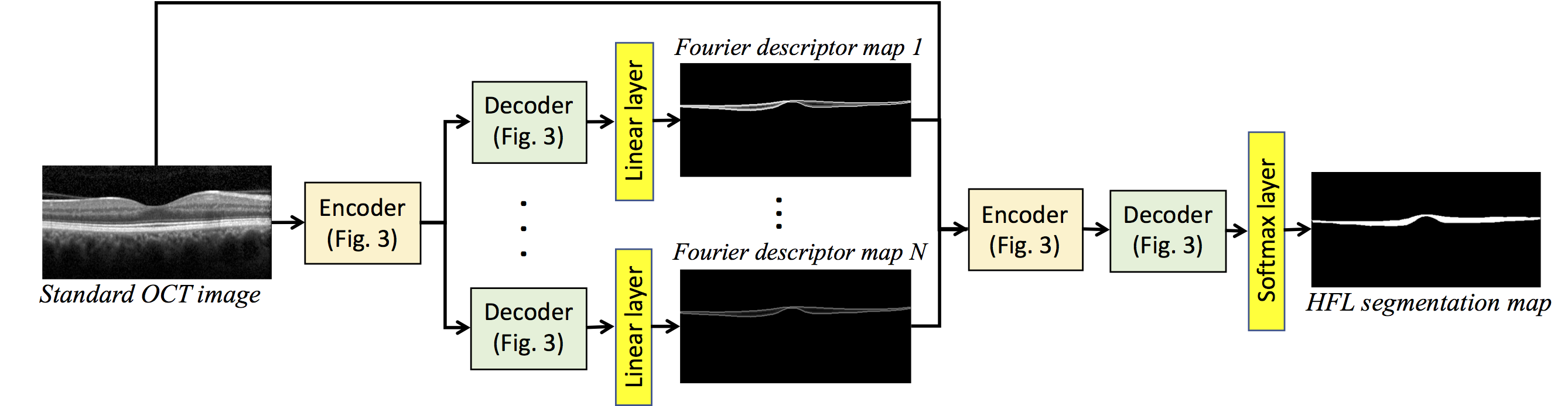}}
\vspace{-0.8cm}
\end{center}
\caption{Architecture of the cascaded FCN to concurrently learn the regression and classification tasks, along with an example standard OCT image as the input and the estimated Fourier descriptor maps and the segmentation map as the outputs of the regression and classification tasks, respectively.}
\label{fig:architecture}
\end{figure*}

\subsection{Fourier Descriptor Map Generation} 
\label{sec:map}

The calculation explained above outputs a set of Fourier descriptors $\texttt{FD}(\gamma_h) = [A_1, A_2, ..., A_N]$ for the contour (outermost pixels) of a single HFL. In order to define the maps of the intermediate regression tasks, which are used to reconstruct a segmentation map, these contour-wise descriptors are mapped onto every pixel, both HFL and background pixels using an iterative algorithm. This algorithm starts with calculating the Fourier descriptors on the contour of HFL in an image and assigns these descriptors to every pixel of the corresponding contour. It then shrinks HFL removing these pixels and repeats the same procedure for the new contour of the shrunk HFL. The algorithm iteratively continues until there exists no HFL contour in the image. The default value of 0 is used as the descriptors of the background pixels. Note that since $N$ Fourier descriptors are calculated for a given contour, this algorithm creates $N$ maps for the image pixels. 

For an example OCT image, Fig.~\ref{fig:maps} illustrates the map of the first Fourier descriptors calculated with respect to the given HFL annotations. Note that this calculation is possible only for a training image whose HFL is manually annotated. For test images, for which annotations are not available, the maps are to be estimated by the intermediate regression tasks of the cascaded FCN.

\subsection{Cascaded Network Architecture and Training}
\label{sec:fcn}

The proposed \emph{FourierNet} model uses a cascaded FCN to concurrently learn the regression and classification tasks. This cascaded FCN uses a multi-task network to estimate $N$ Fourier descriptor maps from an input image in its initial (intermediate) regression tasks. This multi-task network has one encoder to learn shared feature maps and $N$ decoders, each of which learns a different Fourier descriptor map from the shared features. The selected architectures of the encoder and decoders are similar to those of the U-Net model~\cite{ronneberger2015u}. These architectures are illustrated in Fig.~\ref{fig:base-model}. Then, it consecutively uses another U-Net that consists of an encoder and a decoder with the same architecture (Fig.~\ref{fig:base-model}) to predict segmentation labels from the estimated Fourier descriptor maps along with the input image in its final classification task.

In all these selected architectures, the convolutional layers use $3 \times 3$ filters and are followed by the rectified linear unit (ReLU) activation function except the output layers. The output layers of the regression and classification tasks use the linear and softmax activations, respectively. The pooling and upsampling layers use $2 \times 2$ filters. Long-skip connections are added between the corresponding layers of the encoder and the decoder. Dropout layers with the rate of 0.2 are used for regularization. The number of layers and the feature maps used in each convolution layer are illustrated in Fig.~\ref{fig:base-model}. 

The overall architecture of the cascaded FCN is also illustrated in Fig.~\ref{fig:architecture}. This cascaded design, which obtains the Fourier descriptor representation in the middle of the entire architecture, enables both the encoder-decoder networks to exploit the feature maps learned for this representation through gradients flowing in the entire network. The cascaded FCN is implemented in Python using the Keras deep learning library. It is end-to-end trained from scratch to concurrently learn the intermediate regression and the final classification tasks, for which the mean square error and the categorical cross-entropy are used as the loss function, respectively. All tasks contribute to the joint loss function equally with the unit weight. The AdaDelta optimizer is used to adaptively adjust the learning rate and the momentum. The batch size is selected as 1 and early stopping is used in the training.

\section{Experiments}

\subsection{Dataset}

The \emph{FourierNet} model was tested on a dataset that contains the standard OCT scans of 30 eyes belonging to 17 healthy subjects. Eyes were imaged using the Heidelberg SD-OCT imaging equipment (Spectralis$\textsuperscript{\textregistered}$, Heidelberg Engineering GmbH, Heidelberg, Germany) with the standard protocol, in which the incident angle of the light beam was directed to the foveal center. For each eye, there is a set of 49 grayscale OCT scans.  An input image with the resolution of $256 \times 512$ pixels was cropped from each of these scans. Manual annotation of HFL was performed by two ophthalmologists (M.H. and C.K.) on the directional OCT scans, in which OCT images were acquired with superior, inferior, nasal, and temporal tilts in addition to the standard imaging. It is important to note that the OCT images acquired with superior, inferior, nasal, and temporal tilts were used only for annotating HFL but not for training the \emph{FourierNet} model.

The dataset is divided into training and test sets containing the standard OCT scans of 20 and 10 eyes, respectively. The training set is further split into training images, on which the network weights are learned by backpropagation, and validation images, which are used for early stopping. The network training stops if there is no improvement on the validation set loss in the last 50 epochs. Test images are used for final evaluation of the model but not used in any step of the training. The training set contains 735 OCT images of 15 eyes (49 images from each eye), the validation set contains 245 OCT images of five eyes, and the test set contains 490 OCT images of 10 eyes.

\subsection{Comparisons}

The proposed \emph{FourierNet} model is compared with three other algorithms that also use the same encoder and decoder architectures given in Fig.~\ref{fig:base-model}. The \emph{standard-OCT-Unet} algorithm takes a standard OCT image as its input and outputs an HFL segmentation map. It uses neither Fourier descriptors to quantify the HFL shape nor a cascaded design to learn these Fourier descriptors. This is the baseline algorithm. 

The \emph{directional-OCT-Unet} algorithm takes images acquired by the directional OCT as its inputs and also outputs an HFL segmentation map without using a cascaded design. In addition to the standard OCT image, this algorithm also uses the four other images acquired with superior, inferior, nasal, and temporal tilts. This comparison algorithm is used in order to understand a target performance when the directional OCT is available, which is indeed not the case in common practice. This study aims to achieve this performance using only the standard OCT image as an input.

The \emph{directional-OCT-cascaded} algorithm employs the images of the directional OCT in a cascaded network design similar to the one given in Fig.~\ref{fig:architecture}. Similar to \emph{FourierNet}, this algorithm takes a standard OCT image as its input and formulates HFL segmentation as concurrent learning of regression and classification tasks. In its regression task, this algorithm estimates the four other images acquired with superior, inferior, nasal, and temporal tilts from the standard OCT image using a multi-task network. In its classification task, it predicts an HFL segmentation map from these four estimated  images along with the standard OCT image, which is the original input. This comparison algorithm is used to understand the effectiveness of using an intermediate task that estimates the Fourier descriptors in a cascaded FCN design.

\subsection{Evaluation}

\begin{figure}
\centering
\small{
\begin{tabular}{cc}
\includegraphics[width=0.46\columnwidth]{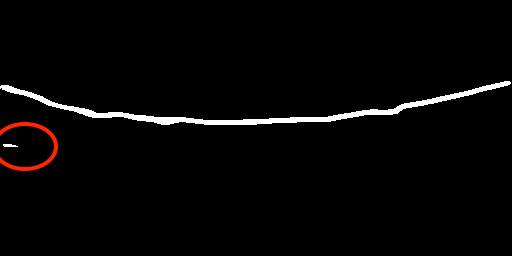} &
\includegraphics[width=0.46\columnwidth]{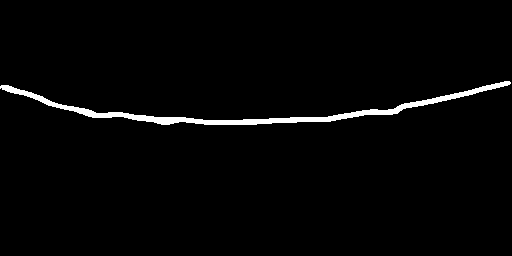} \vspace{-0.07cm} \\ 
(a) & (b) \\
\includegraphics[width=0.46\columnwidth]{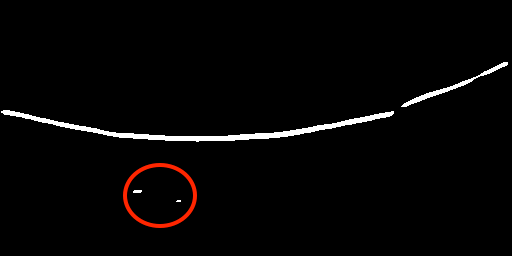} &
\includegraphics[width=0.46\columnwidth]{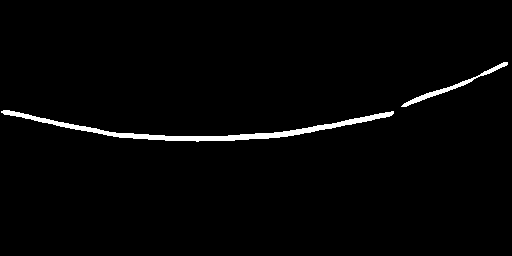} \vspace{-0.07cm}\\ 
(c) & (d)
\end{tabular}
}
\caption{(a)-(b) Segmentation map estimated by the \emph{FourierNet} model and the resulting map after postprocessing. (c)-(d) Segmentation map estimated by the standard-OCT-Unet comparison algorithm and the resulting map after postprocessing. The red ovals indicate noisy regions.}
\label{fig:postprocessing}
\end{figure}

For a test image, pixels whose estimated class posteriors are greater than 0.5 are considered as HFL pixels. Some images may contain small noisy regions that are incorrectly classified as HFL (see Figs.~\ref{fig:postprocessing}a and~\ref{fig:postprocessing}c). These noisy regions are observed in only a few images and they only slightly affect the quantitative results. However, in order to carry out volumetric and thickness analyses of HFL in Sec.~\ref{sec:etdrs}, these regions are eliminated applying a very simple postprocessing method. For each column in the image, this method checks whether the column contains pixels of different connected components. If it does, the method keeps only the pixels of the largest component in the corresponding column, eliminating those of the others (see Figs.~\ref{fig:postprocessing}b and~\ref{fig:postprocessing}d). This postprocessing method is applied on the outputs of the \emph{FourierNet} model as well as the comparison algorithms. 

The resulting HFL maps are evaluated visually and quantitatively on the test set images. For each image, we first find the number of true positive (TP), false positive (FP), and false negative (FN) pixels, comparing the estimated segmentation maps with their manual annotations, and then calculate the precision = TP / (TP + FP), recall = TP / (TP + FN), and f-score metrics. These metrics are averaged over the test set images. Furthermore, for the proposed \emph{FourierNet} model as well as the comparison algorithms, the networks are trained three times and the average metrics of these three runs together with their standard deviations are reported. 

\section{Results}

The quantitative test set results obtained by the algorithms are reported in Table~\ref{table:results}. This table reveals that \emph{FourierNet} achieves better performance compared to the \emph{standard-OCT-Unet} algorithm, which is the baseline. However, it is worth to noting that even this baseline algorithm gives a reasonable performance on the task of HFL segmentation. This suggests that deep learning has a great potential to analyze OCT images in more detail than the common practice. Moreover, the proposed \emph{FourierNet} model, which uses only the standard OCT scan as its input, gives the f-score metric as high as the \emph{directional-OCT-Unet} algorithm (even slightly higher), which uses four more OCT scans acquired with superior, inferior, nasal, and temporal tilts in addition to the standard one as its inputs. This indicates the effectiveness of representing the prior knowledge on the HFL shape in a cascaded network design. 
\begin{table}[t]
\caption{Test set results obtained by the proposed \emph{FourierNet} model and the comparison algorithms. These are the average test set results of the three runs and their standard deviations.}
\begin{center}
\begin{tabular}{ |@{~}p{3.1cm}@{~}|@{~}c@{~}|@{~}c@{~}|@{~}c@{~}| }
 \hline					& {Precision} 			& {Recall} 			& {F-score}		\\  \hline
\emph{FourierNet}  			& 84.96 $\pm$ 1.08		& {\bf 86.17 $\pm$ 0.54}	& {\bf 85.28 $\pm$ 0.32}	\\  \hline
Directional-OCT-Unet 		& {\bf 86.84 $\pm$ 0.40}	& 83.58 $\pm$ 0.68 		& 84.94 $\pm$ 0.35	\\  \hline
Standard-OCT-Unet  		& 86.34 $\pm$ 0.88		& 81.31 $\pm$ 2.03 		& 83.50 $\pm$ 0.68	\\  \hline
Directional-OCT-cascaded 	& 88.43 $\pm$ 0.73		& 79.51 $\pm$ 1.98 		& 83.45 $\pm$ 0.84	\\  \hline
\end{tabular}
\end{center}
\label{table:results}
\end{table}
 
\begin{figure*}
\begin{center}
\begin{tabular}{ll}
\small{Manual HFL annotations} & \small{Manual HFL annotations} \vspace{-0.07cm}\\
\includegraphics[width =0.90\columnwidth]{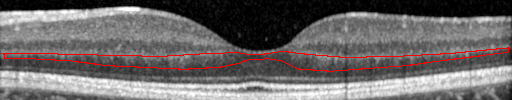} &
\includegraphics[width =0.90\columnwidth]{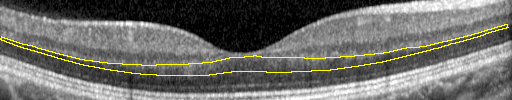} \\ 
\small{{FourierNet}} & \small{{FourierNet}} \vspace{-0.07cm}\\
\includegraphics[width =0.90\columnwidth]{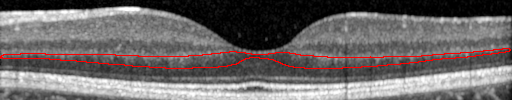} &
\includegraphics[width =0.90\columnwidth]{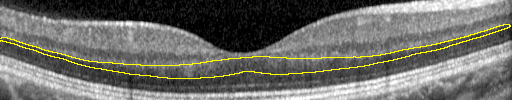} \\ 
\small{{Directional-OCT-Unet}} & \small{{Directional-OCT-Unet}} \vspace{-0.07cm}\\
\includegraphics[width =0.90\columnwidth]{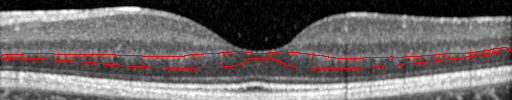} &
\includegraphics[width =0.90\columnwidth]{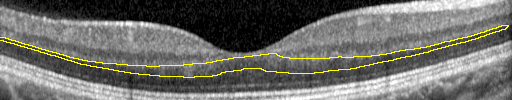} \\
\small{{Standard-OCT-Unet}} & \small{{Standard-OCT-Unet}} \vspace{-0.07cm}\\
\includegraphics[width =0.90\columnwidth]{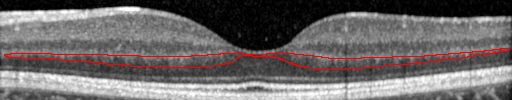} &
\includegraphics[width =0.90\columnwidth]{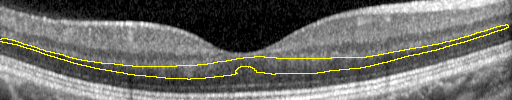} \\
\small{{Directional-OCT-cascaded}} & \small{{Directional-OCT-cascaded}} \vspace{-0.07cm}\\
\includegraphics[width =0.90\columnwidth]{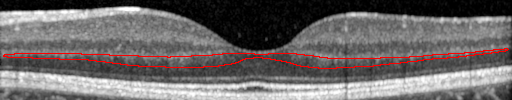} &
\includegraphics[width =0.90\columnwidth]{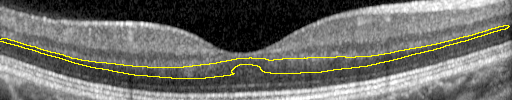} \\
\small{Manual HFL annotations} & \small{Manual HFL annotations} \vspace{-0.07cm}\\
\includegraphics[width =0.90\columnwidth]{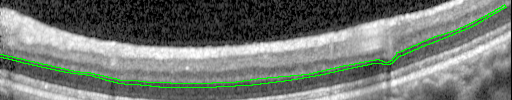} &
\includegraphics[width =0.90\columnwidth]{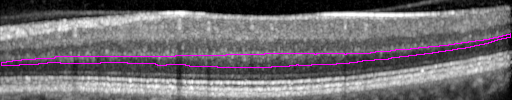} \\ 
\small{{FourierNet}} & \small{{FourierNet}} \vspace{-0.07cm}\\
\includegraphics[width =0.90\columnwidth]{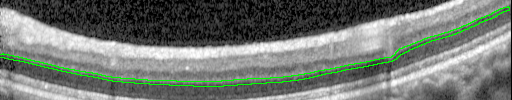} &
\includegraphics[width =0.90\columnwidth]{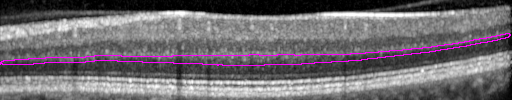} \\ 
\small{{Directional-OCT-Unet}} & \small{{Directional-OCT-Unet}} \vspace{-0.07cm}\\
\includegraphics[width =0.90\columnwidth]{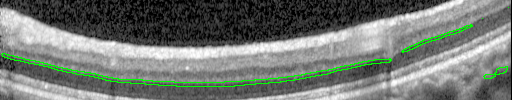} &
\includegraphics[width =0.90\columnwidth]{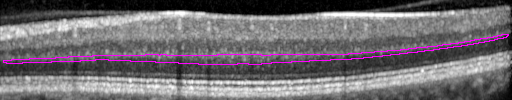} \\ 
\small{{Standard-OCT-Unet}} & \small{{Standard-OCT-Unet}} \vspace{-0.07cm}\\
\includegraphics[width =0.90\columnwidth]{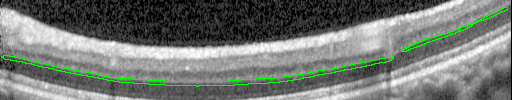} &
\includegraphics[width =0.90\columnwidth]{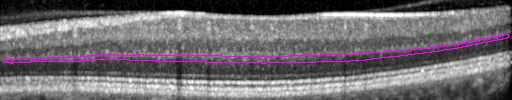} \\
\small{{Directional-OCT-cascaded}} & \small{{Directional-OCT-cascaded}} \vspace{-0.07cm}\\
\includegraphics[width =0.90\columnwidth]{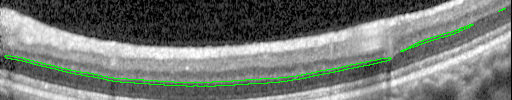} &
\includegraphics[width =0.90\columnwidth]{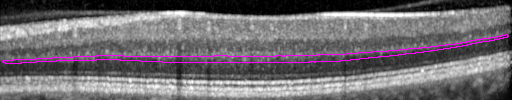} \\ 
\end{tabular}
\end{center}
\caption{Visual results obtained on four exemplary test images. The manual HFL annotations and the results of the \emph{FourierNet} model and the comparison algorithms. All results are embedded on the standard OCT images. The same color is used to show the manual annotations and the results obtained on the same OCT image. }
\label{fig:results}
\end{figure*}

The \emph{directional-OCT-cascaded} algorithm also uses a similar cascaded design but with different intermediate regression tasks. Table~\ref{table:results} also reveals that this way of defining an intermediate task is less effective than using the intermediate task of Fourier descriptor estimation. This may be attributed to the following. Estimating an image from another one may require more complex models (e.g., generative adversarial networks) for this particular task. In contrast, the Fourier descriptor representation, which summarizes the shape prior on HFL, facilitates defining a more effective and more compact learning task. The use of more complex models and the co-use of different tasks in the same cascaded network design are considered as future research direction of this work. The visual results obtained on four exemplary test images are shown in Fig.~\ref{fig:results}. They are also consistent with the quantitative results. 

In our  experiments, we observe that the proposed \emph{FourierNet} model leads to enhanced performance at the fovea (see the center zone of the first two OCT images, for which segmentations are marked with red and yellow, in Fig.~\ref{fig:results}) as well as at the outer regions. It also tends to generate continuous segmentation maps of HFL compared to its counterparts (see the third OCT image, for which segmentations are marked with green, in Fig.~\ref{fig:results}). Additionally, compared to the comparison algorithms, the HFL thickness in the estimated maps of the \emph{FourierNet} model is more consistent with the HFL thickness calculated with reference to the manual HFL annotations. It is important to note that since the HFL thickness is commonly associated with the retina's macular condition, its more correct estimation is important in the common practice. This issue will further be discussed in Sec.~\ref{sec:etdrs}.

\subsection{Parameter Analysis}
\begin{figure}
\begin{center}
\centerline{\includegraphics[width=0.7\columnwidth]{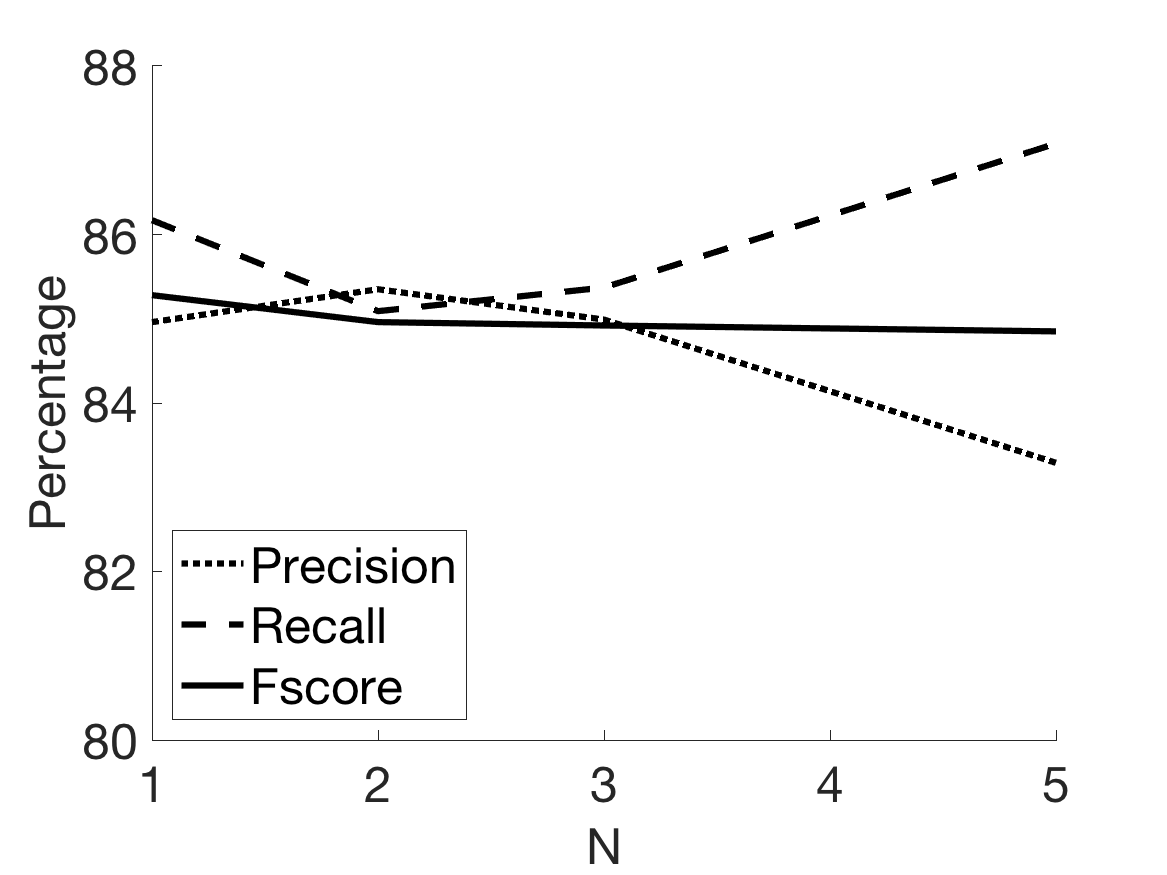}}
\vspace{-0.8cm}
\end{center}
\caption{Test set recall, precision f-scores metrics as a function of the number $N$ of Fourier descriptors. These are the average results obtained over three runs.}
\label{fig:parameter}
\end{figure}

\emph{FourierNet}  has one external parameter $N$, which is the number of Fourier descriptors. In other words, $N$ is the number of Fourier coefficients in the truncated expansion of the distance-to-center function $\xi(l_x)$ that is defined to quantify the HFL contours. First Fourier descriptors contain more general information about this function whereas latter ones contain its finer details. Fig.~\ref{fig:parameter} shows the test set performance as a function of $N$. This figure reveals that the first Fourier descriptor carries sufficient information to capture the HFL contour shape and the latter dimensions do not bring about additional information. Thus, in order to reduce the complexity, $N$ is set to 1 in our experiments. For $N = 1$, the multi-task network given in Fig.~\ref{fig:architecture} has only one decoder, which converts it to a single-task U-Net architecture. Although the first Fourier descriptor is adequate for this particular application of HFL segmentation, this paper presents a more generic algorithm that allows using the latter Fourier descriptors, if this is beneficial for other applications. This possibility can be investigated as future work.

\subsection{ETDRS Grid Analysis}
\label{sec:etdrs}

\begin{figure*}
\begin{center}
\centerline{\includegraphics[width=13cm]{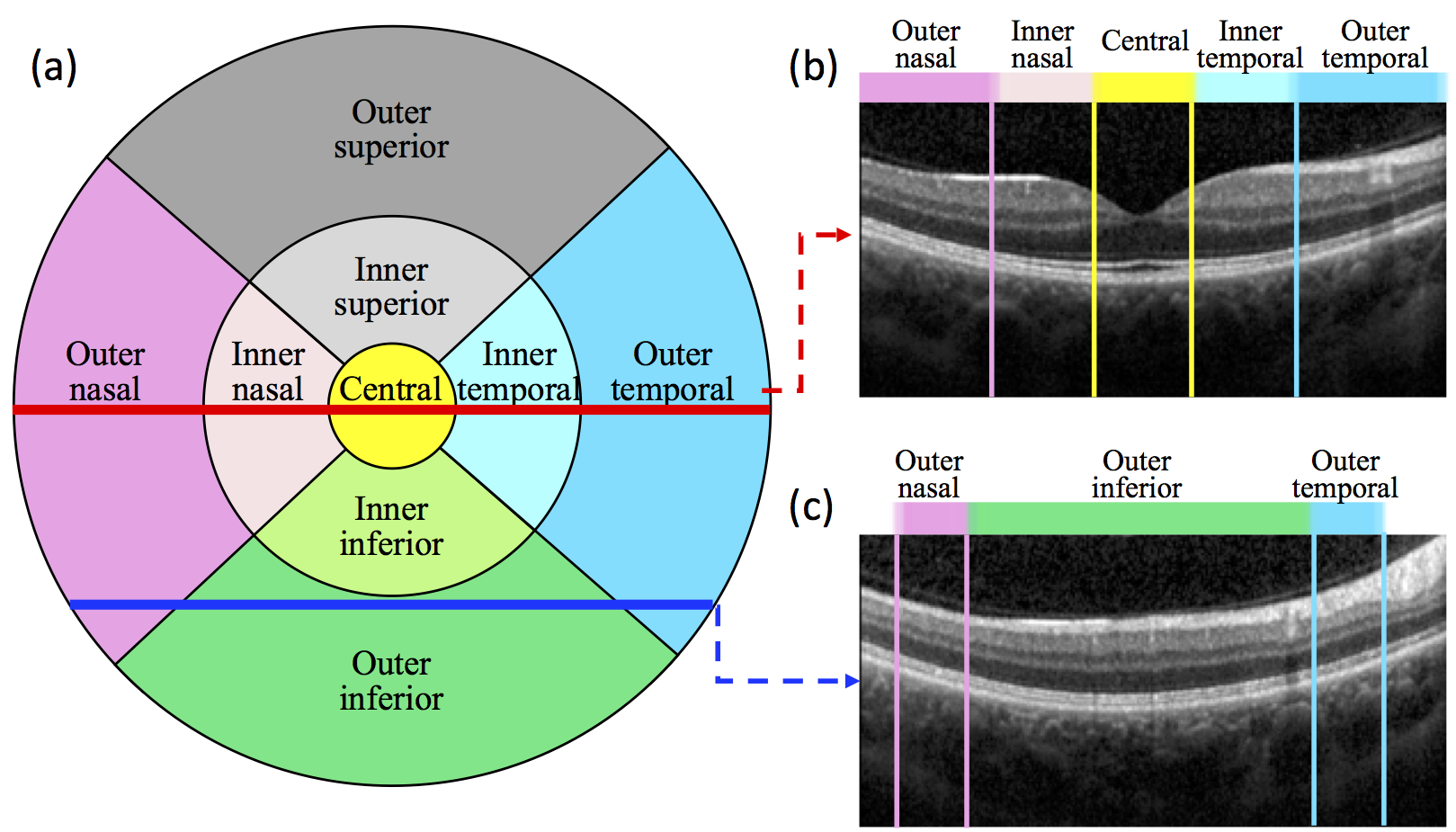}}
\vspace{-0.8cm}
\end{center}
\caption{(a) Illustration of the ETDRS grid that divides the retina volume into nine macular sectors based on the three concentric circles with 1mm, 3mm, and 6mm diameters. This illustration represents the top view for a retinal OCT volume scan. This ETDRS grid is depicted for the right eye; the nasal and temporal sectors should be the opposite for the left eye. (b)-(c) Two samples of cross-sectional OCT scan slices corresponding to the red and blue lines in the illustrated ETDRS grid.}
\label{fig:etdrs}
\end{figure*}

In evaluating the macular condition of a retina, it is very common to use the Early Treatment of Diabetic Retinopathy Study (ETDRS) grid that divides the retina into nine macular sectors based on the three concentric circles with 1mm, 3mm, and 6mm diameters (see Fig.~\ref{fig:etdrs}). It might be necessary to analyze the layer's volume and thickness in the whole ETDRS grid as well as in each of these sectors for evaluating the retina's macular condition. In order to understand the effects of using the proposed \emph{FourierNet} model on this ETDRS grid analysis, we carry out additional experiments.

First, the precision, recall, and f-score metrics are calculated for each of the nine sectors of the ETDRS grid separately by considering only the pixels falling in the corresponding sector. The sector-based average f-score metrics obtained on the test set together with their standard deviation are reported in Table~\ref{table:etdrs-results}. This table demonstrates that \emph{FourierNet} achieves better f-scores in all of the nine sectors than the comparison algorithms that also use only the standard OCT image as their inputs. Furthermore, compared to the directional-OCT-Unet algorithm, which uses four more OCT scans acquired with superior, inferior, nasal, and temporal tilts in addition to the standard one as its input, \emph{FourierNet} leads to sometimes slightly better sometimes slightly worse f-scores but without using any additional OCT scans. 
\begin{table*}[t]
\caption{Test set results obtained by the \emph{FourierNet} model and the comparison algorithms. These are the sector-based average test set f-scores of the three runs and their standard deviations. See Fig.~\ref{fig:etdrs} for how the sectors are determined.}
\vspace{-0.3cm}
\begin{center}
\begin{tabular}{ |@{~}p{3.1cm}@{~}|@{~}c@{~~}|@{~~}c@{~~}|@{~~}c@{~~}|@{~~}c@{~~}|@{~~}c@{~~}|@{~~}c@{~~}|@{~~}c@{~~}|@{~~}c@{~~}|@{~~}c@{~}| }
  \hline                                             &                 		& {Inner}		& {Inner}		& {Inner}           & {Inner}           & {Outer}           & {Outer}          & {Outer}          & {Outer}      \\ 
                                                         & {Central}      	    	& {superior}	& {inferior}       	& {nasal}          & {temporal}    & {superior}      & {inferior}       & {nasal}          & {temporal}  \\  \hline
\emph{FourierNet}                       & 85.8 $\pm$ 0.9 & {\bf 90.0 $\pm$} 0.3 & 89.4 $\pm$ 0.8 & {\bf 89.6 $\pm$ 0.7} & 89.8 $\pm$ 0.2 & {\bf 86.9 $\pm$ 0.3} & 84.0 $\pm$ 0.7 & {\bf 84.5 $\pm$ 0.4} & {\bf 84.2 $\pm$ 0.3}\\  \hline
Directional-OCT-Unet                 & {\bf 86.1 $\pm$ 1.2} & 89.5 $\pm$ 0.7 & {\bf 90.4 $\pm$ 1.0} & {\bf 89.6 $\pm$ 0.5} & {\bf 90.1 $\pm$ 0.3} & 85.8 $\pm$ 0.3 & {\bf 85.2 $\pm$ 0.6} & 83.5 $\pm$ 0.5 & 83.0 $\pm$ 0.3\\  \hline
Standard-OCT-Unet                    & 84.4 $\pm$ 1.6 & 88.0 $\pm$ 1.3 & 88.7 $\pm$ 0.8 & 89.0 $\pm$ 0.5 & 88.0 $\pm$ 0.6 & 85.3 $\pm$ 0.8 & 81.5 $\pm$ 0.7 & 83.6 $\pm$ 0.3 & 81.8 $\pm$ 1.4\\  \hline
Directional-OCT-cascaded 	& 81.9 $\pm$ 0.7 & 87.0 $\pm$ 1.2 & 87.2 $\pm$ 1.9 & 87.8 $\pm$ 1.4 & 87.8 $\pm$ 1.6 & 86.0 $\pm$ 0.7 & 83.0 $\pm$ 1.6 & 82.7 $\pm$ 1.2 & 82.5 $\pm$ 0.4\\  \hline
\end{tabular}
\end{center}
\label{table:etdrs-results}
\end{table*}

Next, we conduct volumetric and thickness analyses of HFL within the ETDRS grid. For that, the total volume and the average thickness of HFL are calculated with reference to manual annotations as well as with reference to a segmentation map obtained by the \emph{FourierNet} model as well as each of the comparison algorithms. The average values calculated over three runs are reported in Table~\ref{table:etdrs-volume-thickness}. This table reveals that more accurate volume and thickness estimations are possible with the proposed \emph{FourierNet} model, which indicates the effectiveness of representing the shape prior of HFL in a cascaded network design. 
\begin{table}[t]
\caption{Volumetric and thickness data of HFL within the ETDRS grid. These are calculated on the test set images with reference to the manual annotations as well as the segmentation maps of the algorithms. The total volume is given in $\text{mm}^3$ and the average thickness is given in $\mu\text{m}$.}
\begin{center}
\begin{tabular}{ |@{~}p{3.1cm}@{~}|c|c| }
 \hline					& {\bf ~~~Total volume~~~} 		& {\bf Average thickness}		\\  \hline
Manual annotations 			& 0.71 $\pm$ 0.05		& 25.36 $\pm$ 1.63 		\\  \hline
\emph{FourierNet}  			& 0.70 $\pm$ 0.07		& 24.76 $\pm$ 2.44 		\\  \hline
Directional-OCT-Unet 		& 0.67 $\pm$ 0.07		& 23.82 $\pm$ 2.55 		\\  \hline
Standard-OCT-Unet  		& 0.65 $\pm$ 0.06		& 23.14 $\pm$ 2.30 		\\  \hline
Directional-OCT-cascaded 	& 0.61 $\pm$ 0.07		& 21.79 $\pm$ 2.33 		\\  \hline
\end{tabular}
\end{center}
\label{table:etdrs-volume-thickness}
\end{table}

\section{Conclusion}

This paper presents a shape-preserving network for automated HFL segmentation in standard OCT scans of the retina. This network, which we call \emph{FourierNet}, relies on benefiting the shape prior of HFL in its training. To this end, it proposes to quantify the shape prior by extracting Fourier descriptors on the HFL contours and introduces a new cascaded network design that learns these descriptors concurrently with the segmentation task. We tested our \emph{FourierNet} model for HFL segmentation on 1470 OCT images of 30 eyes. Our experiments revealed that \emph{FourierNet} achieved segmentation in scans acquired by standard OCT imaging at least with the performance that would be obtained when directional OCT imaging was used. As a result, this model proved to be a useful tool for HFL segmentation by reducing the necessity to perform directional OCT imaging.

In our experiments, we obtained promising results on OCT scans of healthy retina. One future research direction is to test the model also on diseased retina in order to associate the volumetric and thickness changes in HFL with the macular condition of the diseased retina. This might require more advanced postprocessing methods due to the structural disorders expected in the diseased retina. This paper used HFL segmentation as a showcase application. Using the proposed shape-preserving network for other segmentation problems is considered as another future research direction of this study.

\end{document}